\documentclass[acmtog,review=false,nonacm]{acmart}
\acmSubmissionID{1682}

\usepackage{booktabs} 

\usepackage[ruled]{algorithm2e} 

\SetAlFnt{\small}
\SetAlCapFnt{\small}
\SetAlCapNameFnt{\small}
\SetAlCapHSkip{0pt}

\usepackage{pifont}

\copyrightyear{2025}
\acmYear{2025}
\setcopyright{cc}
\setcctype{by}
\acmConference[SA Conference Papers '25]{SIGGRAPH Asia 2025 Conference Papers}{December 15--18, 2025}{Hong Kong, Hong Kong}
\acmBooktitle{SIGGRAPH Asia 2025 Conference Papers (SA Conference Papers '25), December 15--18, 2025, Hong Kong, Hong Kong}\acmDOI{10.1145/3757377.3763908}
\acmISBN{979-8-4007-2137-3/2025/12}
\begin{document}
\title{4DSloMo: 4D Reconstruction for High Speed Scene with Asynchronous Capture}

\author{Yutian Chen}
\orcid{1234-5678-9012-3456}
\affiliation{%
 \institution{Shanghai AI Laboratory}
 \city{Shanghai}
 \country{China}}
 \affiliation{%
 \institution{The Chinese University of Hong Kong}
 \city{Hong Kong}
 \country{China}}
\email{chenyt0205@gmail.com}

\author{Shi Guo}
\affiliation{%
 \institution{Shanghai AI Laboratory}
 \city{Shanghai}
 \country{China}}
\email{guoshi@pjlab.org.cn}

\author{Tianshuo Yang}
\affiliation{%
\institution{The University of Hong Kong}
\city{Hong Kong}
\country{China}}
\email{yangtianshuo@connect.hku.hk}

\author{Lihe Ding}
\affiliation{%
 \institution{The Chinese University of Hong Kong}
 \city{Hong Kong}
 \country{China}}
\email{dean.dinglihe@outlook.com}

\author{Xiuyuan Yu}
\affiliation{%
 \institution{The Chinese University of Hong Kong}
 \city{Hong Kong}
 \country{China}}
\email{1155211255@link.cuhk.edu.hk}

\author{Jinwei Gu}
\affiliation{%
 \institution{NVIDIA}
 \city{San Jose}
 \country{USA}}
\email{gujinwei@gmail.com}

\author{Tianfan Xue}
\affiliation{%
 \institution{The Chinese University of Hong Kong}
 \city{Hong Kong}
 \country{China}}
 \affiliation{%
 \institution{Shanghai AI Laboratory}
 \city{Shanghai}
 \country{China}}
\email{tfxue@ie.cuhk.edu.hk}

\begin{abstract}
Reconstructing fast-dynamic scenes from multi-view videos is crucial for high-speed motion analysis and realistic 4D reconstruction. However, the majority of 4D capture systems are limited to frame rates below 30 FPS (frames per second), and a direct 4D reconstruction of high-speed motion from low FPS input may lead to undesirable results. In this work, we propose a high-speed 4D capturing system only using low FPS cameras, through novel capturing and processing modules. On the capturing side, we propose an asynchronous capture scheme that increases the effective frame rate by staggering the start times of cameras. By grouping cameras and leveraging a base frame rate of 25 FPS, our method achieves an equivalent frame rate of 100–200 FPS without requiring specialized high-speed cameras. On processing side, we also propose a novel generative model to fix artifacts caused by 4D sparse-view reconstruction, as asynchrony reduces the number of viewpoints at each timestamp. Specifically, we propose to train a video-diffusion-based artifact-fix model for sparse 4D reconstruction, which refines missing details, maintains temporal consistency, and improves overall reconstruction quality. Experimental results demonstrate that our method significantly enhances high-speed 4D reconstruction compared to synchronous capture. Project page: \href{https://openimaginglab.github.io/4DSloMo/}{https://openimaginglab.github.io/4DSloMo/}
\end{abstract}

%
%


%


\begin{teaserfigure}
   \centering
   \includegraphics[width=1\textwidth]{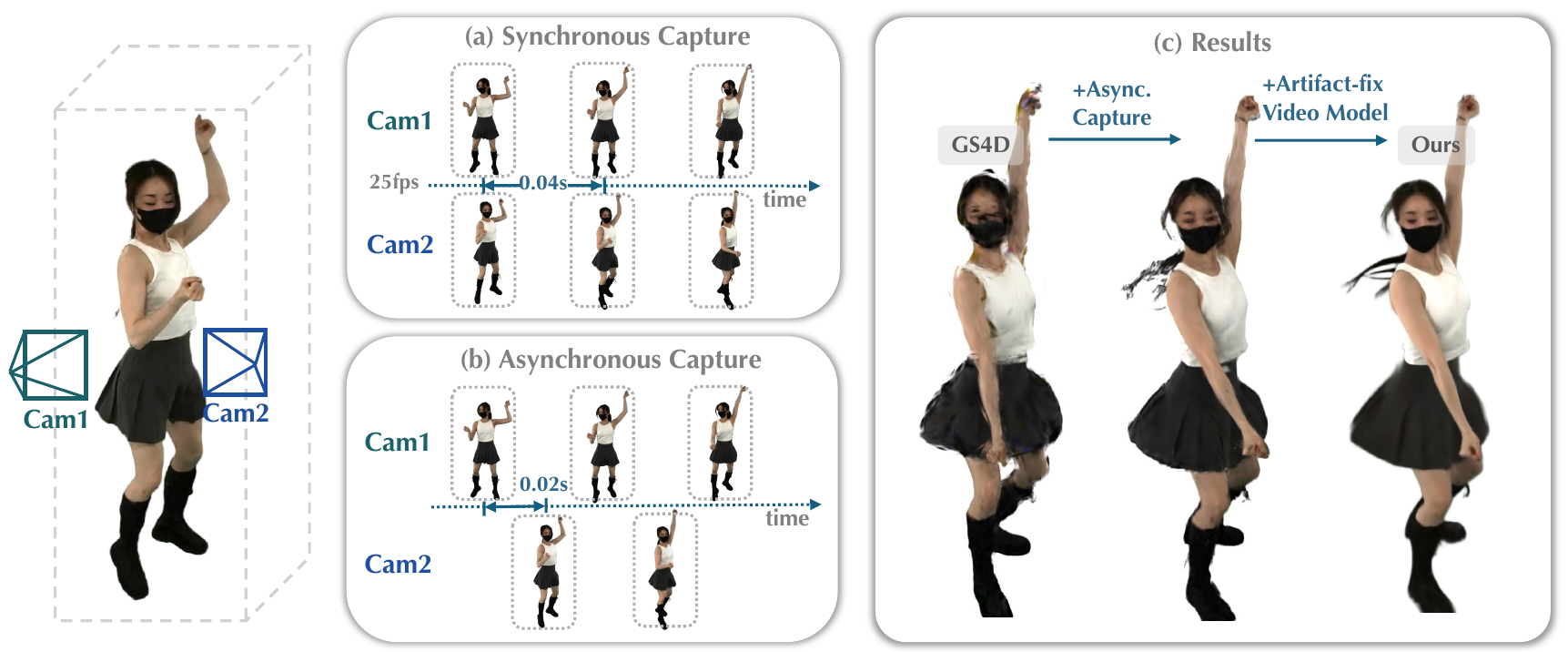}
   \caption{\textbf{Our 4D Reconstruction Results of the real-captured scene.} We propose an asynchronous capture scheme, which increases the effective capture frame rate by staggering the start times of cameras without any additional cost. We further leverage video diffusion priors to enhance the reconstruction results. The results show that our method can reconstruct high speed and complex motion with high quality.}
   \label{fig:teaser}
\end{teaserfigure}

\maketitle

\section{Introduction}
\label{sec:intro}
Fast-dynamic scenes reconstruction from multi-view videos is a fundamental challenge in 3D vision with broad applications. In sports and biomechanics, high-speed 4D reconstruction enables precise motion capture for athlete performance evaluation and injury prevention. In autonomous driving and robotics, accurately reconstructing 3D models of fast-moving objects, such as pedestrians and vehicles, is essential for perception and decision-making. Additionally, in VR/AR content production, reconstructing high-fidelity human performances, such as dance and martial arts, is crucial for creating realistic digital avatars.

Still, 4D reconstruction of fast-moving objects remains challenging, as the majority of 4D-capturing camera arrays operate at no more than 30 FPS (frames per second). For instance, DNA-Rendering \cite{2023dnarendering} operates at 15 FPS, while ENeRF-Outdoor~\cite{lin2022efficient} and Neural3DV~\cite{li2022neural} achieve 30 FPS. In contrast, many activities, such as cloth movement shown in Fig.~\ref{fig:teaser}, require a higher frame rate; for example, standard high-speed 2D photography typically operates at 120 FPS or higher to capture them. Extending frame rate of existing camera arrays is hard, as that requires more expensive and dedicated hardware, and greatly increases data transmission bandwidth requirements.

Another way to capture high-speed 4D motion without hardware changes is to increase the frame rate at the reconstruction stage. Recently, gaussian-splatting-based 4D reconstruction methods~\cite{li2022neural,lin2024gaussian,wu20244d,wang2024shape,wang2025vidu4d,xu20244k4d} have significantly improved novel view synthesis for dynamic scenes. For simple motion, it can reconstruct continuous frames from sparse temporal inputs, effectively increasing the frame rate. However, it still fails to handle complex nonlinear motion, such as cloth movement, resulting in obvious artifacts, as shown in the top right of Fig.~\ref{fig:teaser}. Thus, this raises an interesting question: is it possible to recover intermediate frames of high-speed motion using regular video cameras at 30 FPS?

To achieve that, we propose a novel asynchronous capture scheme to increase frame rate. We deliberately add delay to the starting time between different cameras, so different cameras can capture different timestamps, effectively boosting the frame rate. An example is shown in Fig.~\ref{fig:teaser}. Two synchronized cameras can only achieve 25 FPS (top row) capturing. However, in our setup, by differing the capture time of \textit{Cam2}, the capture interval reduces to 0.02s, achieving 50FPS capturing. In practice, we use eight 25FPS cameras, and divide them into 4 or 8 groups to effectively increase the perceived temporal resolution to 100 or 200 FPS, respectively. By capturing temporally denser frames, we can more accurately model the intermediate motion information, particularly for complex motion.


A key challenge introduced by such asynchronous capturing is the limited number of available views at each timestamp, which increases viewpoint sparsity and results in visible artifacts in the reconstruction, as illustrated in Fig.~\ref{fig:teaser}. Recent 3D sparse view reconstruction methods~\cite{yu2024lm,wynn2023diffusionerf,gao2024cat3d,wu2024reconfusion} utilized the pre-trained image diffusion model to prove the reconstruction results. However, we experimentally find that utilizing image diffusion model to perform artifacts-fix cause temporal inconsistency for sparse 4D reconstruction problem, due to frame-wise independent refine. Thus we propose to train the video-diffusion based artifact-fix model for 4D reconstruction. To train this model, we construct a dataset by temporally sub-sampling 4D sequences to simulate large motion and our proposed asynchronous capturing pattern. The sparsely sampled sequences are used to train a 4D Gaussian Splatting model, whose rendered outputs naturally exhibit reconstruction artifacts. These degraded outputs are then paired with the original ground-truth sequences to form training data. Despite the limited size of available 4D datasets, our artifact-fix video diffusion model demonstrates strong generalization across multiple scenes, benefiting from the powerful spatiotemporal priors inherent in video diffusion model. Notably, with only 750 training pairs, our method effectively removes reconstruction artifacts and significantly improves visual quality, as shown in Fig.~\ref{fig:teaser}.

We compare our method with several state-of-the-art approaches \cite{Wu_2024_CVPR,yang2023gs4d,kplanes_2023} using the DNA-Rendering dataset~\cite{cheng2023dna} and the Neural4DV dataset~\cite{li2022neural}. To simulate fast motion and asynchronous capture, we apply temporal subsampling. To further validate our approach on real-world asynchronous multi-view video data, we capture 12 sequences of fast and complex dynamic scenes using an asynchronous capture setup. Experimental results demonstrate that our method improves high-speed 4D reconstruction compared to synchronized capture.

Our key contributions are summarized as follows:

\begin{itemize}
    \item \textbf{Hardware solution: } We design and implement a low-cost asynchronous capture system that increases the effective frame rate  by staggering the start times of commodity cameras.
    \item \textbf{Software solution: }We train an artifact-fix video diffusion model to refine the 4D reconstruction results, significantly improving the visual quality of the rendered images.
    \item \textbf{Dataset}: We develop the first dataset containing 12 sequences of asynchronous multi-view videos to validate our approach, demonstrating the practical feasibility and effectiveness of our method.
\end{itemize}

\begin{figure*}[!t]
   \centering
   \includegraphics[width=1\textwidth]{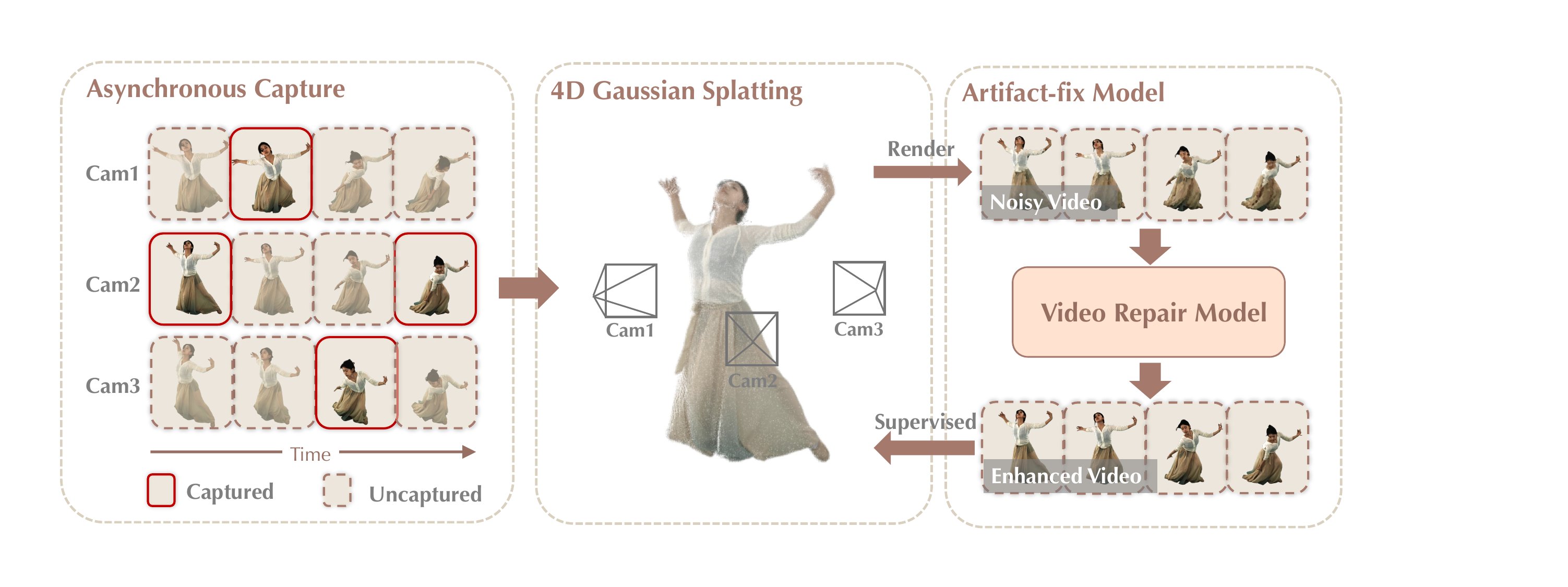}
   \caption{\textbf{The overall pipeline of our model.} Given several asynchronous multi-view videos, we first initialize a 4D Gaussian model for a specific iteration. We then employ an artifact-fix video diffusion model to refine the input videos. The refined videos are subsequently used to update the 4D Gaussian model.}
   \label{fig:pipeline}
\end{figure*}

\section{Related Work}
\label{sec:formatting}


\subsection{Dynamic 3D Reconstruction}
Reconstruction and novel view synthesis of dynamic 3D scenes remain challenging. 
Many existing methods require multiple synchronized videos captured from different viewpoints as input \cite{kplanes_2023,lin2024gaussian,xu20234k4d,yang2023gs4d,lin2022efficient,Wu_2024_CVPR}. These works tend to use radiance field models such as NeRF~\cite{mildenhall2021nerf} or 3DGS~\cite{kerbl20233d} as the underlying static 3D representations, and model motion patterns using various types of the deformation field, for example, MLPs~\cite{pumarola2021d, yang2024deformable,lu20243d}, spatial-temporal planes~\cite{Wu_2024_CVPR,kplanes_2023}, polynomial functions~\cite{li2024spacetime} and Fourier series~\cite{katsumata2024compact}. These methods learn motion patterns from independent time input in a vanilla way and not adept at frames interpolation in the temporal dimension. 

Existing camera arrays for 4D scene capture are often limited to low frame rates (e.g., 15-30 FPS [cheng2023dna, lin2022efficient, li2022neural]), posing challenges for reconstructing high-speed dynamics. While specialized systems like multi-view photometric stereo~\cite{vlasic2009dynamic} offer higher quality, they typically require complex lighting and are costly.

To improve temporal resolution without specialized hardware, one direction is to manipulate the camera exposure process. Wu et al.~\cite{wu2012performance} pioneered staggering the exposure of multiple cameras to increase the effective frame rate, which then requires model-based deblurring in post-processing. A similar principle was applied to a single camera via coded exposure to embed high-speed information into one frame~[liu2013efficient]. Other works like TimeFormer~[jiang2024timeformer] enhance temporal modeling purely through algorithms, but they lack the new physical information needed to reconstruct truly rapid motion.

Our work adopts asynchronous capture but replaces model-based deblurring with a video-diffusion model. This generative approach corrects sparse-view artifacts to refine detail and ensure temporal consistency, removing the need for a pre-defined geometric proxy.

\subsection{Sparse View Scene Reconstruction}

As the asynchronous capture method reduces the number of available views at each timestamp, our work is closely related to the field of sparse view scene reconstruction. Sparse view reconstruction is a challenging problem due to the limited availability of input views, which can lead to incomplete or ambiguous scene representations. To address this challenge, several recent works have proposed innovative solutions. For example, Freenerf~\cite{yang2023freenerf} introduced a method that incorporates depth regularization to improve the reconstruction quality from sparse views. Similarly, Nerf in 3d vision~\cite{gao2022nerf} and  Regnerf~\cite{niemeyer2022regnerf} explored frequency regularization techniques to enhance the robustness and accuracy of scene reconstruction when only a few input views are available.

In addition to regularization-based methods, several recent works have leveraged extra supervision signals to guide the reconstruction process. For instance, SPARF~\cite{sparf2023} utilized optical flow estimated from pre-trained models to provide additional constraints during optimization. Monosdf~\cite{yu2022monosdf} integrates both depth and normal maps as supplementary supervision signals. This combination helps refine the surface details of the reconstructed scene, enhancing its geometric accuracy. Därf~\cite{song2023darf} further explored the use of depth maps as an auxiliary signal in their method. Collectively, these works demonstrate the potential of combining various forms of supervision and regularization to enhance sparse view reconstruction.

\subsection{3D/4D Reconstruction with Diffusion Model}
The recent progress in diffusion models has driven significant advancements in 3D and 4D applications. 
Pioneer work ReconFusion~\cite{wu2024reconfusion} trains a novel view synthesis diffusion model conditioning on sparse input views and then adopts a SDS (Score Distillation Sampling)~\cite{poole2022dreamfusion} style optimization strategy for novel view supervision. Unlike the SDS loss, ReconFusion directly predicts the pseudo ground truth of novel views by sampling from the trained diffusion model at each optimization step, which is then used to compute the reconstruction loss.
Follow-up works~\cite{yu2024lm,wynn2023diffusionerf,gao2024cat3d} also integrate diffusion models with NeRF or 3DGS. To further improve the optimization efficiency, Deceptive-NeRF~\cite{liu2023deceptive} and 3D-GS Enhancer first render pseudo images from the sparse-view-reconstructed 3D representation and use a diffusion model to enhance these pseudo views to obtain high-quality novel view supervision without querying the diffusion model at every optimization step. Inspired by these few-shot 3D reconstruction methods, we utilize a diffusion model to remove the artifacts from the 4D model rendered images.

\section{Method}

To achieve high-quality 4D reconstruction for scenes with large and complex motion, we first revisit the data acquisition process and propose a novel asynchronous capture scheme (Sec.~\ref{sec:method-asy-capture}) that surpasses the conventional 30 FPS (frame-per-second) limitation. Although the captured data can be leveraged by state-of-the-art method, \emph{i.e.}, GS4D~\cite{yang2023gs4d} (revisited in Sec.~\ref{sec:method-4D-pre}), to achieve temporally dense 4D reconstruction, the asynchronous capture scheme inherently introduces challenges of sparse viewpoints and temporal inconsistency. To address the artifacts caused by sparse views, a artifact-fix video diffusion model is introduced in Sec.~\ref{sec:method-artifact-fix-diffusion}, followed by the 4D reconstruction process with diffusion priors in Sec.~\ref{sec:method-4d-recon-diffusion}. The overall framework of our approach is illustrated in Fig.~\ref{fig:pipeline}.

\subsection{Preliminary: 4D Gaussian Splatting}
\label{sec:method-4D-pre}
We choose 4D Gaussian Splatting (GS4D)~\cite{yang2023gs4d} as the 4D representation for reconstruction. In GS4D, dynamic 3D scenes are represented by introducing an additional time dimension $t$ into anisotropic 3D Gaussians~\cite{kerbl20233d}. Formally, a 4D Gaussian is defined by its mean vector $\mu \in \mathbb{R}^4$ and covariance matrix $\Sigma \in \mathbb{R}^{4 \times 4}$:
\begin{equation}
    p(\mathbf{x} | \mu, \Sigma) = \exp \left( -\frac{1}{2} (\mathbf{x} - \mu)^T \Sigma^{-1} (\mathbf{x} - \mu) \right),
\end{equation}
where $\mathbf{x} = (x, y, z, t)$, with $(x,y,z)$ represents the spatial coordinates, and $t$ denotes the temporal coordinate. The covariance matrix $\Sigma$ is factorized into a scaling matrix $S$ and a rotation matrix $R$, similar to 3D Gaussian splatting: $\Sigma = R S S^T R^T$, where $S = \text{diag}(s_x, s_y, s_z, s_t)$ defines the anisotropic scaling of the Gaussian, and $R$ is a 4D rotation matrix parameterized by two quaternions $q_l, q_r$, ensuring a valid rotation in the 4D space: $R = L(q_l) R(q_r)$.

The rendering process follows the standard differentiable rasterization used in Gaussian splatting. Given a pixel $(u, v)$ at time $t$, the final rendered image $x_r$ is computed by blending the visible 3D Gaussians:
\begin{equation}
    x_r(u, v, t) = \sum_{i=1}^{N} p_i(u, v, t) \alpha_i c_i(d, t) \prod_{j=1}^{i-1} (1 - p_j(u, v, t) \alpha_j),
\end{equation}
where $p_i(u, v, t)$ is the probability density of the $i$-th Gaussian at pixel $(u, v, t)$, $\alpha_i$ represents its opacity, and $c_i(d, t)$ is the time-dependent color modeled by 4D spherical harmonics (4DSH) ~\cite{yang2023gs4d}. 

\subsection{Asynchronous Capture Scheme}
\label{sec:method-asy-capture}

To capture 4D scenes, given a set of cameras at different viewpoints $\{P_{i}\}_{i=1}^N$, the previous methods \cite{2023dnarendering, li2022neural} would synchronize the capture timing of $N$ cameras to capture images simultaneously, as shown in Fig.~\ref{fig:teaser} (a). However, for high-speed motion, none of this $N$ cameras capture the information between two neighboring consecutive frames. The normal frame rate of cameras is less than or equal to 30FPS and 4D reconstruction may fail for large and complex motion that requires high frame rate~\cite{cheng2023dna,li2022neural,lin2022efficient}.

To capture a temporally denser information, we propose a novel asynchronous capture scheme that enables cameras to start capturing at different time instants, as shown in Fig.~\ref{fig:teaser} (b). Specifically, we put \( K \) cameras into a group (for example, \( K = 2 \) in Fig.~\ref{fig:teaser} (b)), starting at different times for the capture: the \( i \)-th camera starts capturing an image at the timestamp  
$t_i = i \cdot (\tau/K) + j \cdot \tau$,  
where \( i \in \{1, 2, \ldots, K\} \) denotes the index of the camera within the group, and \( j \) represents the frame index in the video sequence recorded at a frame rate of 25 FPS. This design temporally staggers the capture timing of different cameras within the 1/25 second exposure intervals, effectively increases the frame rate of the camera system by a factor of \( K \) without introducing additional hardware costs. In the experiment, we choose $K=4$, which effectively pushes the frame rate to 100 FPS. 

\subsection{Artifact-fix Video Diffusion Model}
\label{sec:method-artifact-fix-diffusion}
%

\begin{figure}[!t]
   \centering
   \includegraphics[width=0.5\textwidth]{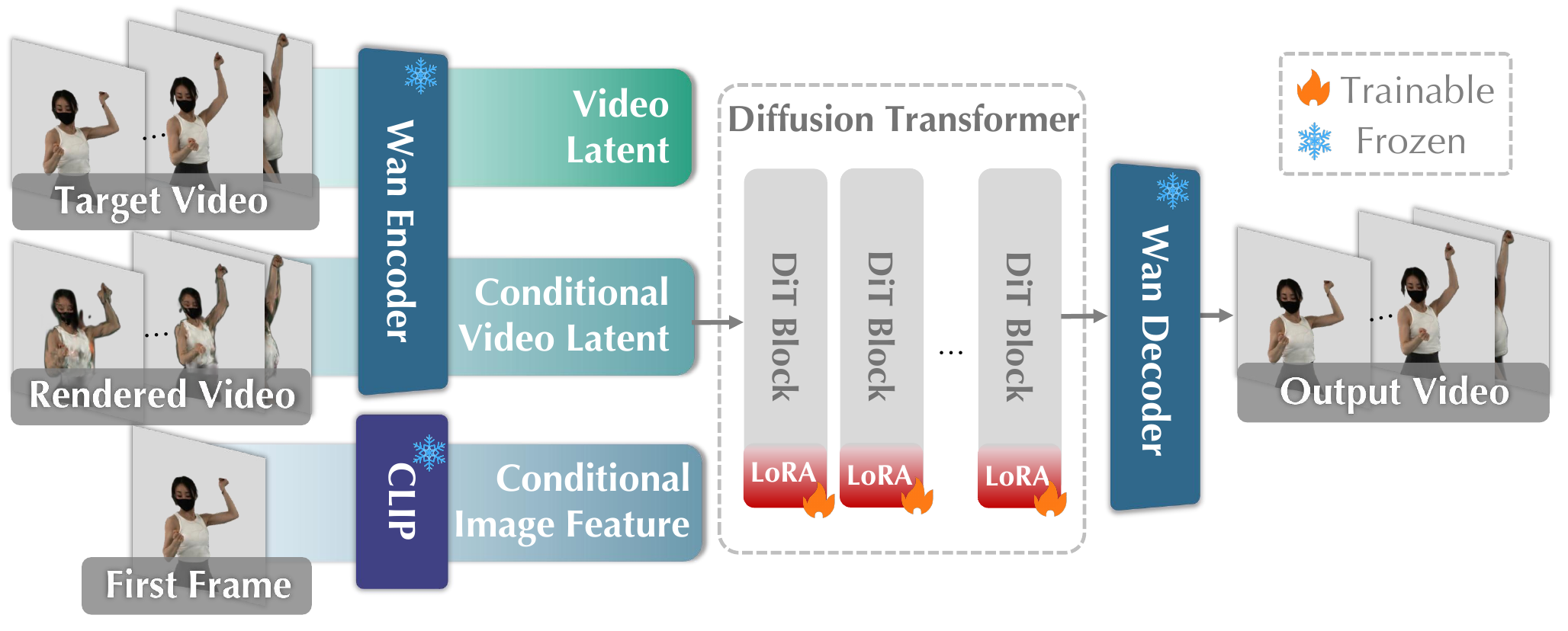}
   \caption{\textbf{Illustration of artifact-fix video diffusion model setup.} Our model freezes all parameters in the network, except for the LoRA weights, to fine-tune a video diffusion model. Precisely, we integrate Lora parameters into the DiT model. With a LoRA rank designated as 16, this integration takes place in each transformer block. }
   \label{fig:wan}
\end{figure}

\begin{figure*}[!t]
    \centering
    \includegraphics[width=1\textwidth]{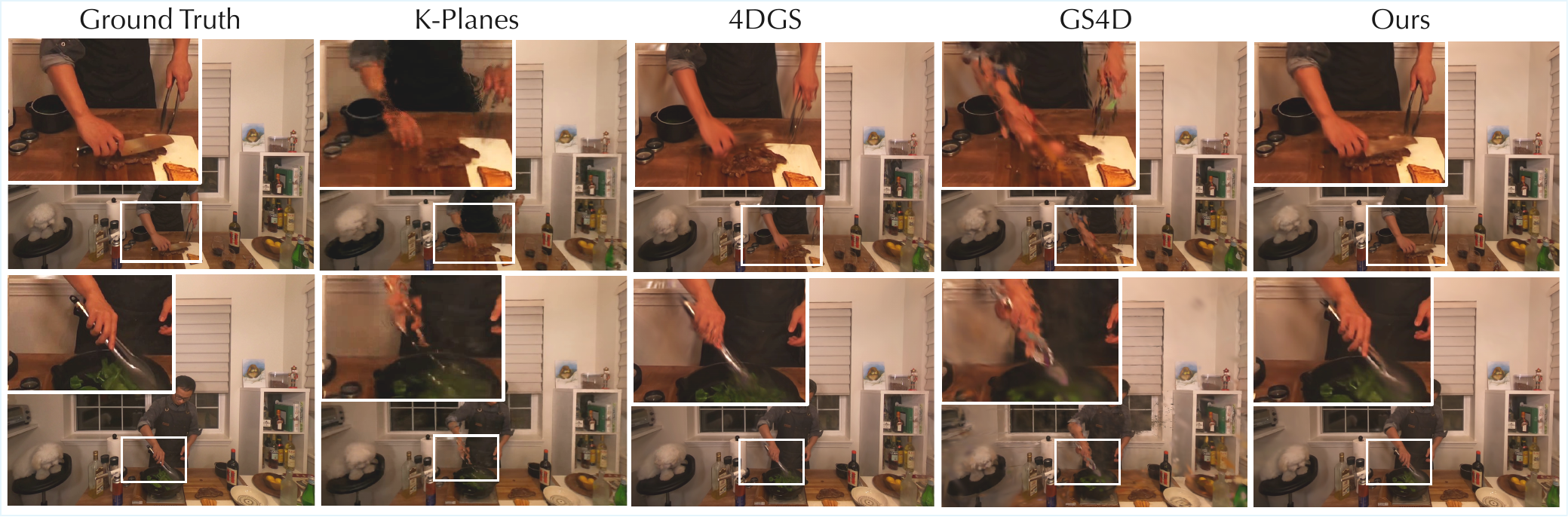}
    \caption{Qualitative result on the Neural3DV dataset~\cite{li2022neural}.}
    \label{fig:compare_n3v2}
\end{figure*}

Although the asynchronous capture introduced in the previous section increase the effective frame rate by $K$ times to better capture highspeed movement, it also introduce additional artifacts. Particularly, at each time stamp, asynchronous capture reduces the number of viewpoints by $K$ times, resulting in ``floater" artifacts in the 4D reconstruction as shown in Fig.~\ref{fig:teaser} (c) due to the sparse view reconstruction challenges. Although there are some recent sparse view 3D reconstruction methods~\cite{yu2024lm,wynn2023diffusionerf,gao2024cat3d,wu2024reconfusion,wu2025difix3d} can leverage image diffusion priors to enhance reconstruction quality, directly applying them to 4D scenes still causes temporal inconsistency due to frame-wise independent processing.

Therefore, we propose a novel artifact-fix video diffusion model to solve this challenge in 4D sparse reconstruction. Particularly, we take a pretrained video diffusion model and fine-tune on 4D reconstruction data to tailor for our tasks.
The model takes the rendered video $V^{render} \in \mathbb{R}^{C \times T \times H \times W}$ containing floater artifacts as input and generates temporally coherent, sharp, and outputs a clean video $\hat{V} \in \mathbb{R}^{C \times T \times H \times W}$. This clean video will later be used to refined 4D Gaussian model, which we will discuss in~Sec.~\ref{sec:method-4d-recon-diffusion}. Below we introduce details of our model.



\textbf{Data curation.} First, 
to train the model, we constructed a dataset consisting of pairs of videos with artifacts and corresponding clean version as ground truth. Specifically, we asynchronously sub-sample the multi-view video along the temporal dimension, as shown in Fig.~\ref{fig:pipeline}(a), to train a 4D Gaussian Splatting (GS4D) model. The trained GS4D is then used to render videos at the original frame rate, producing noisy outputs containing reconstruction artifacts. These rendered videos are paired with the corresponding clean videos to form the training data for diffusion model fine-tuning.

\textbf{Fine-tuning.} We build our artifact-fix model on top of the video diffusion model Wan2.1~\cite{wan2025} as shown in Fig.~\ref{fig:wan}. To be specific, we use Wan-VAE $\mathcal{E}$ to compress rendered video $V^{render}$ and target video ${V^{target}}$ into the latent space $z^{render},z^{target}\in \mathbb{R}^{c \times t \times h \times w}$. The noise latent $z_t^{target}$ and condition latent $z^{render}$ are concatenated along the channel axis and then passed through the DiT model of Wan2.1. 
To repurpose the original image-to-video generation setting in Wan2.1 for video enhancement, we use all frames of $z^{\text{render}}$ as conditioning to guide artifact repair across the entire sequence, as described in~\cite{wan2025}.

During fine-tuning, we freeze the encoder and decoder, and apply LoRA-based fine-tuning only to the DiT component. The training loss is defined as:
\begin{equation}
    \mathcal{L}_{tune} = \mathbb{E}_{z^{target},t,\epsilon,z^{render}}[||(\epsilon_\theta(z_t^{target},t,z^{render},c^{tex})-\epsilon)||_2^2]
\end{equation}
where $c^{tex}$ denotes an object-specific language prompt. 

\subsection{4D Reconstruction with Diffusion Prior}

\label{sec:method-4d-recon-diffusion}
With the artifact-fix video diffusion model $\mathcal{M}$, we can improve the quality of 4D Gaussian model. As shown in Fig.~\ref{fig:pipeline}, we first reconstruct an initial 4D Gaussian models $\mathcal{G}$ from all the captured images. Then, for each training view, we render a high-frame-rate video $V^{render}$, covering all timestamps observed by any of the cameras. Because the initial 4D Gaussian is reconstructed from sparse views, these videos $V^{render}$ contain floater artifacts, but can provide the diffusion model~$\mathcal{M}$ with essential spatial viewpoint information and temporal object motion information. 

To remove those artifact, we then obtain the latent representation \( z^{render} \in \mathbb{R}^{c \times t \times h \times w} \) by applying Wan-VAE to \( V^{render} \), and concatenate it with pure noise of the same shape along the channel dimension. This combined representation is then used to generate the refined video \( \hat{V} \), which is clearer and sharper.

Finally, the refined video \( \hat{V} \) is used to supervise the rendering process through the following loss function:
\begin{equation}
    L_{diff} = || V^{render} - \hat{V}||_1 + L_p(V^{render},\hat{V})
\label{equa:diff}
\end{equation}
where $L_p$ is the perceptual distance LPIPS~\cite{zhang2018unreasonable}. 

The effectiveness of this approach is illustrated in Fig.~\ref{fig:teaser}(c). The proposed asynchronous capture can recover fast-dynamic motion compared to conventional synchronous capture. However, due to the inherently sparse views at each time step, the learned Gaussian representation introduces noticeable artifacts. By applying our per-scene artifact-fix diffusion model to refine the Gaussian representation, these artifacts can be effectively suppressed, leading to improved final results.

\section{Experiments}

\subsection{Implementation Details}

\begin{figure*}[!t]
    \centering
    \includegraphics[width=0.9\textwidth]{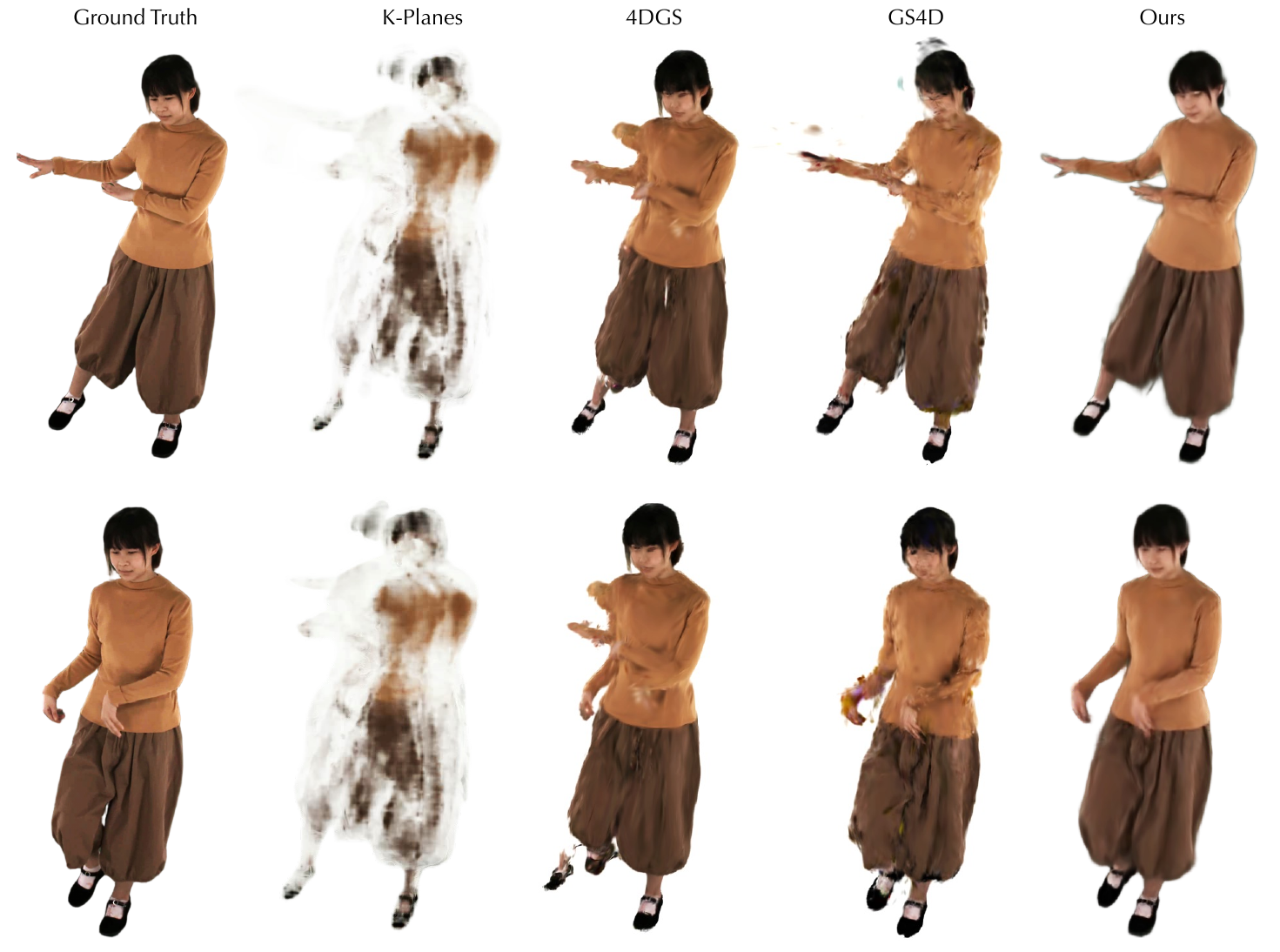}
    \caption{Qualitative result on the DNA-Rendering dataset~\cite{cheng2023dna}.}
    \label{fig:compare_dna}
\end{figure*}

Our framework is illustrated in Fig.~\ref{fig:pipeline}. We adopt 4D Gaussian Splatting (GS4D)~\cite{yang2023gs4d} as the underlying 4D representation. To enhance visual quality, we build an artifact-correction video diffusion model based on Wan-I2V-14B~\cite{wan2025}, fine-tuning its DiT backbone with injected LoRA layers. The model is trained on 750 noisy-clean video pairs from the DNA-Rendering dataset~\cite{cheng2023dna}, using a learning rate of $10^{-4}$ and a LoRA rank of $16$.

Next, we train the 4D Gaussian representation. The 4DGS model undergoes an initial optimization phase for 7k iterations. However, the output of this stage still contains noticeable artifacts. To mitigate these artifacts, we apply the trained artifact-fix diffusion model. In the subsequent 7k iterations, the optimization incorporates only with the refined videos produced by the diffusion model.

\subsection{Datasets}
\begin{figure}[!t]
   \centering
   \includegraphics[width=0.45\textwidth]{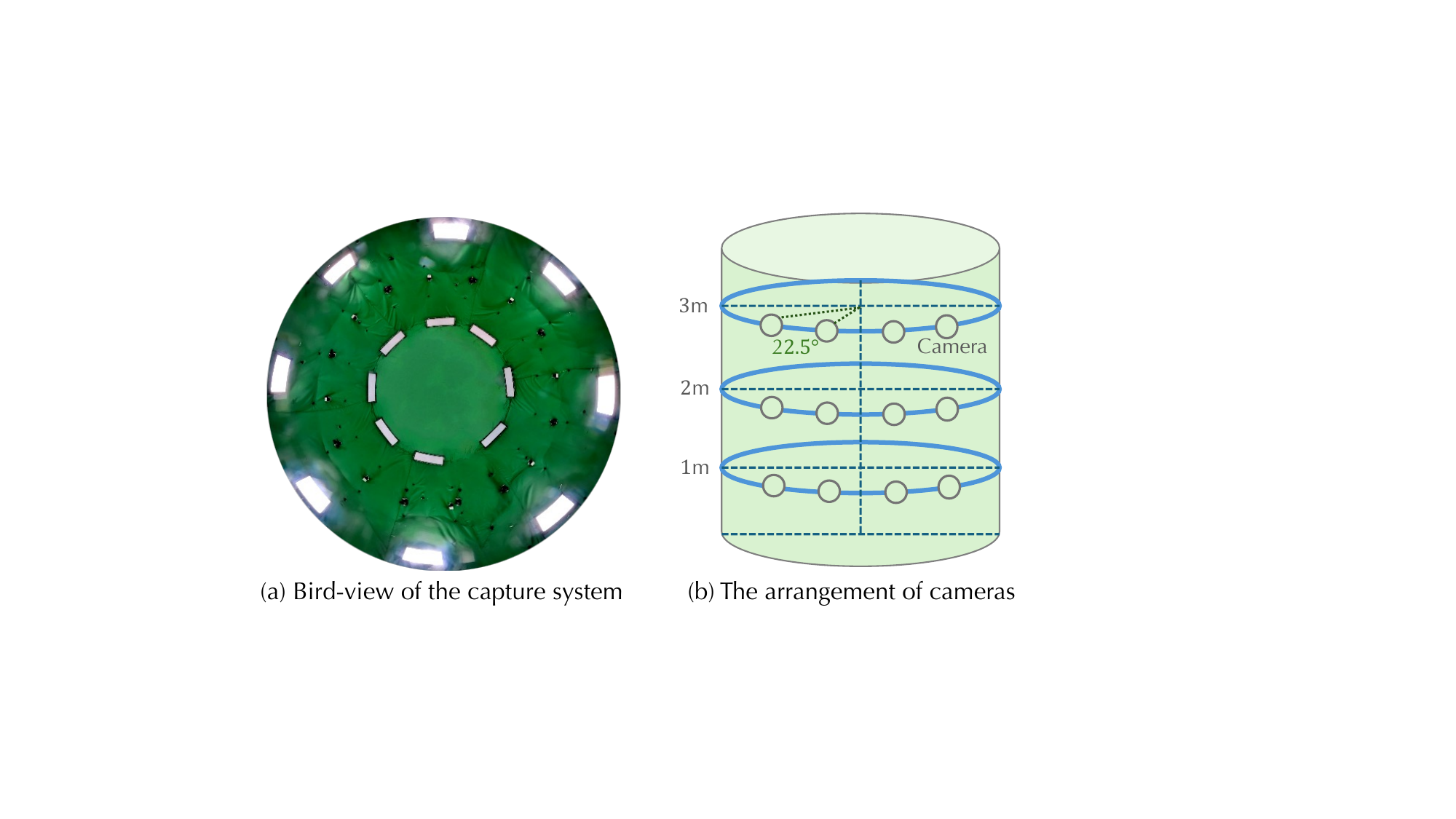}
   \caption{{Illustration of the capture system.}}
   \vspace{-0.2cm}
   \label{fig:camera}
\end{figure}

\begin{figure*}[!t]
   \centering
   \includegraphics[width=0.83\textwidth]{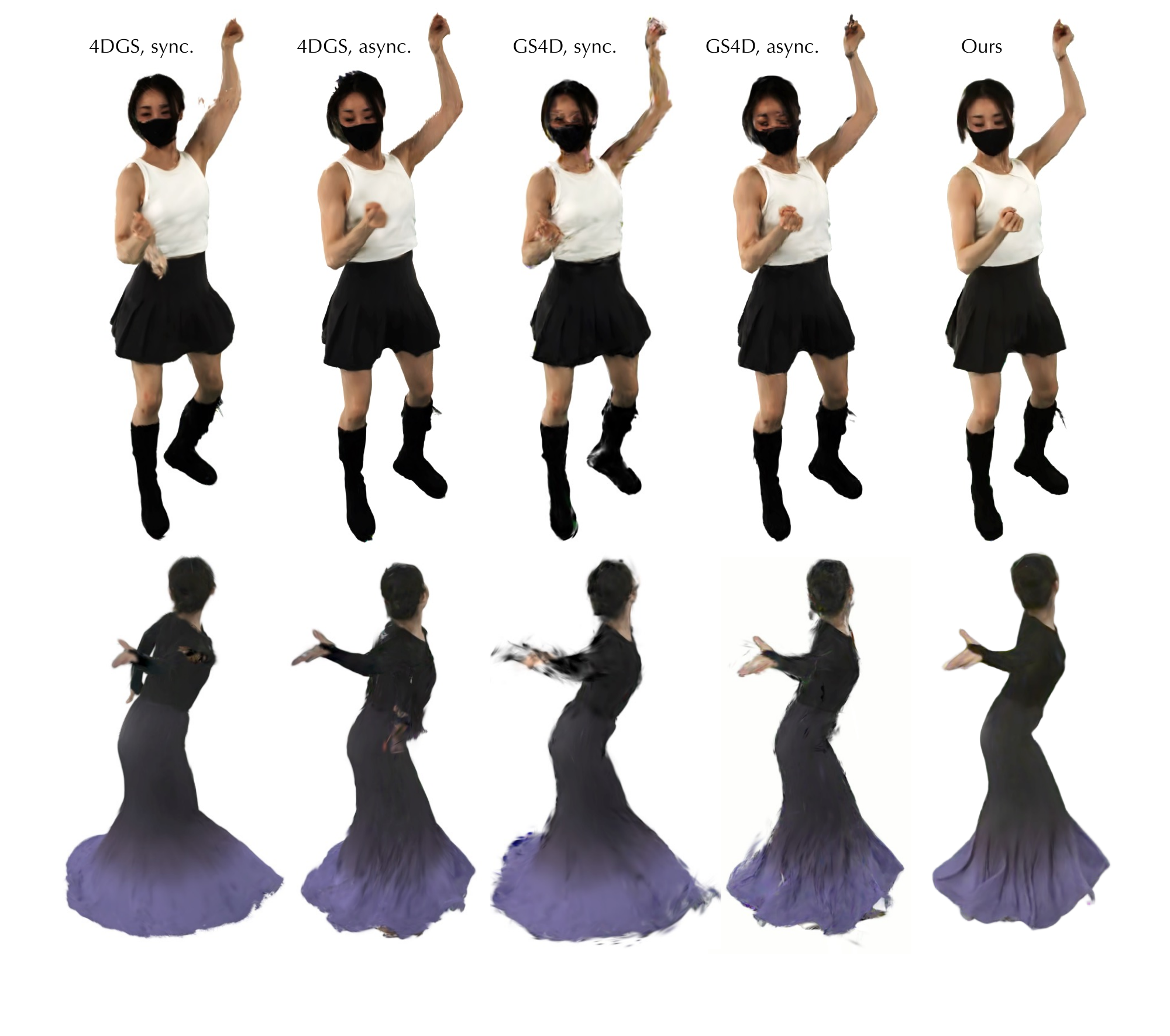}
   \caption{Quality result on our real-capture dataset.}
   \label{fig:real-capture}
\end{figure*}
We evaluate our method on multiple widely used multi-view datasets, including DNA-Rendering~\cite{2023dnarendering} and Neural3DV~\cite{li2022neural}. DNA-Rendering captures 10-second clips of dynamic human subjects and objects at 15 FPS using a combination of 4K and 2K cameras. To induce large motions, we temporally subsample the dataset to one-fourth of its original frame rate. The test set consists of all frames from a held-out view. Neural3DV captures multi-view videos at 30 FPS, each lasting ten seconds. We subsample the videos to one-twelfth of their original frame rate. For each scene, one view is held out for testing, while the remaining views are used for training. By applying temporal downsampling to existing 4D datasets, we effectively simulate large inter-frame motion and generate both synchronous and asynchronous capture settings within the same scene. This setup enables direct comparison between the two acquisition strategies under large-motion conditions.

It is worth noting that the artifact-fix video diffusion model is trained on the DNA-Rendering dataset. We construct a total of 750 noisy-clean video pairs for training, each with a resolution of $1024\times 1024$ and a length of 25 frames.

\textbf{Capturing setup: } Since this is the first work to employ an asynchronous capture strategy for 4D Gaussian Splatting, no existing real-world 4D dataset has been recorded using such a setup. To evaluate our method on real asynchronously captured scenes, we build a custom multi-view camera array for 4D data acquisition. As illustrated in Fig.~\ref{fig:camera}, the setup consists of 12 cameras operating at 25 FPS, all capable of hardware-synchronized triggering. The cameras are arranged at three different vertical levels, with four cameras evenly spaced at each level, approximately 22.5 degrees apart. Although all 12 cameras support synchronous triggering, we manually introduce varying trigger delays across cameras to enable asynchronous capture. By dividing the camera array into four or eight temporal groups, we effectively achieve a capture rate of 100 FPS or 200 FPS, respectively, as discussed in Sec.~\ref{sec:method-asy-capture}.

\begin{figure*}[!t]
   \centering
   \includegraphics[width=0.83\textwidth]{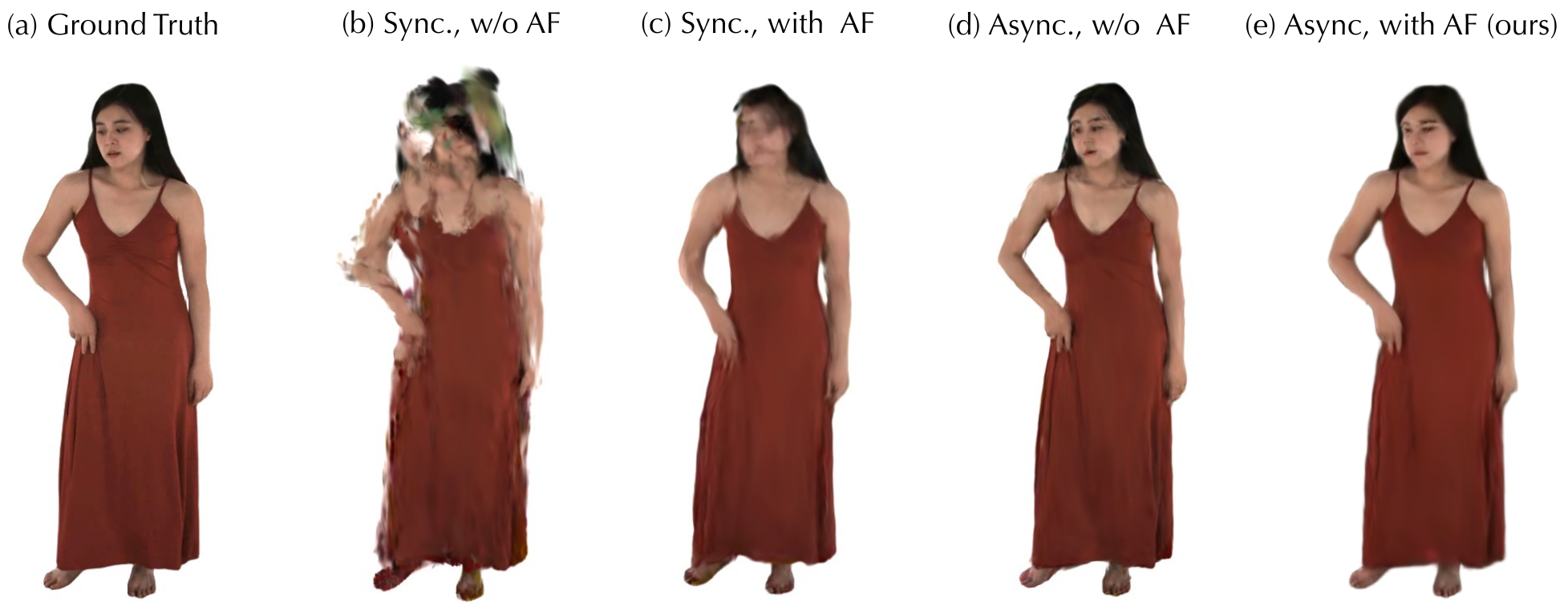}
   \caption{Ablation study of asynchronous capture and diffusion model on the DNA-Rendering dataset~\cite{cheng2023dna}(AF represents artifact-fix video diffusion model).}
   \label{fig:ablation}
\end{figure*}
\textbf{Benchmark for Asynchronous 4D Reconstruction} 
%
Using the above camera array setup, we captured a variety of high-speed motion scenes, including dancing, sports activities, and rapid object interactions such as waving a chess piece. In total, we construct a dataset consisting of 12 sequences of asynchronously recorded multi-view videos, focusing on non-linear and large-motion scenarios. Each video has a resolution of $2048 \times 2248$ pixels.



\subsection{Comparison Experiments}


\begin{table}[t] \setlength{\abovecaptionskip}{0.2cm}
\caption{\textbf{Quantitative comparison on the DNA-Rendering dataset~\cite{2023dnarendering}.} Higher values indicate better performance for metrics marked with ($\uparrow$), while lower values are preferable for metrics marked with ($\downarrow$).}
    \centering
    {
    \begin{tabular}{lccccc}
    
    \hline
    \textbf{Method}  & \textbf{PSNR} $\uparrow$ &\textbf{SSIM } $\uparrow$& \textbf{LPIPS} $\downarrow$\\ \hline
    K-Planes & 22.74&0.750 &0.443 \\
      4DGS \cite{Wu_2024_CVPR}&23.09 &0.772 &0.351  \\
      GS4D \cite{yang2023gs4d}&24.75 & 0.797&0.337  \\ \hline
      Ours & \textbf{26.76} & \textbf{0.845} & \textbf{0.293} \\
 \hline
    \end{tabular}}
    
     \label{tab:result-dna}
\end{table} 

\begin{table}[t] \setlength{\abovecaptionskip}{0.2cm}
\caption{\textbf{Quantitative comparison on the Neural3DV dataset~\cite{li2022neural}.} Higher values indicate better performance for metrics marked with ($\uparrow$), while lower values are preferable for metrics marked with ($\downarrow$).}
    \centering
    {
    \begin{tabular}{lccccc}
    
    \hline
    \textbf{Method}  & \textbf{PSNR} $\uparrow$ &\textbf{SSIM } $\uparrow$ & \textbf{LPIPS} $\downarrow$ \\ \hline
    K-Planes &28.31& 0.875& 0.211\\       
           4DGS \cite{Wu_2024_CVPR}&29.99 & 0.928& 0.252 \\
      GS4D \cite{yang2023gs4d}&30.54 &0.917 & 0.178 \\ \hline
      Ours & \textbf{33.48}& \textbf{0.951}& \textbf{0.134}\\
 \hline
    \end{tabular}}
    
     \label{tab:result-n3v}
\end{table} 

For assessment, we use three metrics, encompassing peak-signal-to-noise ratio (PSNR), structural similarity index (SSIM) and perceptual quality measure (LPIPS). For baselines, we choose three state-of-the-art 4D reconstruction methods, including K-Planes~\cite{kplanes_2023}, 4DGS~\cite{Wu_2024_CVPR} and GS4D~\cite{yang2023gs4d}.

We first conduct a quantitative evaluate on two existing datasets, DNA-Rendering and Neural3DV, as shown in Table \ref{tab:result-dna} and Table \ref{tab:result-n3v}. Our method achieves the best results across fidelity metrics (PSNR and SSIM) as well as perceptual metrics (LPIPS). This demonstrates the effectiveness of asynchronous capture and artifact-fix model for high-frame-rate 4D reconstruction in large-motion scenarios.

\begin{figure*}[!t]
   \centering
   \includegraphics[width=1\textwidth]{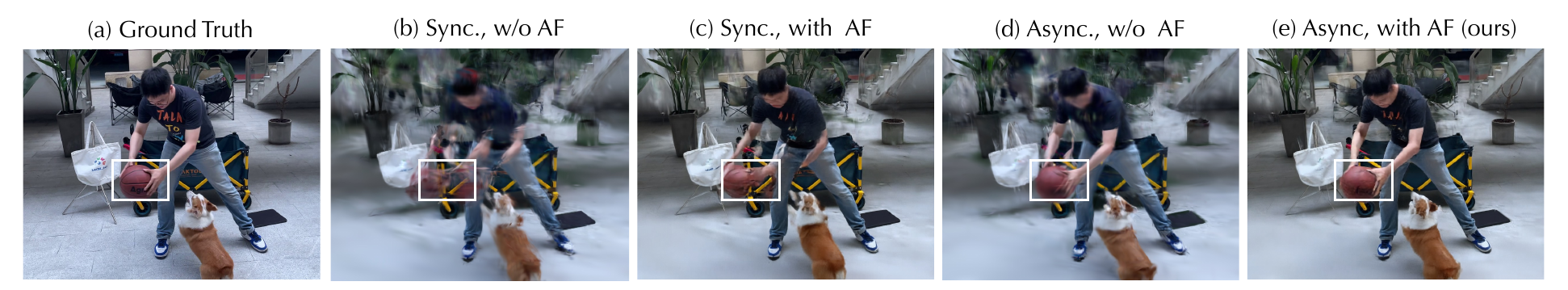}
   \caption{Ablation study of asynchronous capture and diffusion model on LongVolCap datasets~\cite{xu2024longvolcap}(AF represents artifact-fix video diffusion model).}
   \label{fig:ablation2}
\end{figure*}

We also present a quantitative comparisons on Neural3DV dataset~\cite{li2022neural} and DNA-Rendering dataset~\cite{2023dnarendering} in Fig.~\ref{fig:compare_n3v2} and Fig.~\ref{fig:compare_dna} respectively. The results show that our methods can produce higher fidelity renderings using the same size of input data. It becomes apparent that, due to insufficient input information in the temporal dimension, all comparison methods have generated severe artifacts and motion distortions, especially in regions of rapid motion. Specifically, methods like 4DGS and K-Planes, which estimate a deformation field on top of a static canonical representation, tend to become stuck at certain poses when lacking temporal supervision. In contrast, methods like GS4D, which treat temporal and spatial dimensions as an integrated whole, are prone to generating numerous artifacts in the absence of temporal supervision. Supporting this, the experimental results also prove that existing 4D representations are unable to automatically interpolate the temporal dimension to recover plausible motion information. Benefiting from the proposed asynchronous capture, our system hardware-wise obtains sufficient temporal supervision, enabling the reconstruction of plausible motion even in large-motion scenarios, as shown in Fig.~\ref{fig:compare_n3v2} and Fig.~\ref{fig:compare_dna}. Furthermore, leveraging the intrinsic modeling capabilities of the 4D representation in the spatial domain and the diffusion prior, our method effectively mitigates artifacts caused by sparse views, achieving improved temporal-spatial consistency.

The qualitative results on our real-captured dataset is shown in Fig.~\ref{fig:real-capture}. Since we are unable to capture both synchronous and asynchronous videos from the same perspective at the same moment, we captured two separate sequences while maintain the same motion patterns to ensure a fair comparison. The results show that, whether using 4DGS or GS4D, the asynchronous capture method helps the 4D reconstruction model recover more accurate geometry compared to the synchronous method. With the addition of the artifact-fix model, our approach is further capable of reconstructing fast-moving regions (e.g., arms), non-linear motions, and complex texture areas (e.g., skirts) with high quality.

\section{Discussion}

\begin{table}[t] \setlength{\abovecaptionskip}{0.2cm}
\caption{\textbf{Ablation study on DNA-Rendering dataset~\cite{2023dnarendering}.} Higher values indicate better performance for metrics marked with ($\uparrow$), while lower values are preferable for metrics marked with ($\downarrow$).}
    \centering
    {
    \begin{tabular}{lccccc}
    
    \hline
    \textbf{Capture}  & \textbf{Artifact-fix }  &\textbf{PSNR} $\uparrow$ &\textbf{SSIM } $\uparrow$ & \textbf{LPIPS} $\downarrow$ \\ \hline
    Sync. &\ding{55}&24.75 & 0.797&0.337 \\       
           Sync.&\ding{51}&24.15&0.800&0.331  \\
      Async.&\ding{55}&26.23& 0.831&0.315 \\ \hline
      Async. &\ding{51}& \textbf{26.76} & \textbf{0.845} & \textbf{0.293}\\
 \hline
    \end{tabular}}
    
     \label{tab:ablation}
\end{table} 

\subsection{Effect of Asynchronous Capture} 
%

To evaluate the effectiveness of asynchronous capture, we compare it with synchronous capture on the DNA-Rendering dataset and LongVolCap datasets~\cite{xu2024longvolcap}. Quantitative results are shown in Table~\ref{tab:ablation}, and qualitative comparisons are provided in Fig.~\ref{fig:ablation} and Fig.~\ref{fig:ablation2}. Under the original synchronous capture setup, the camera system’s frame rate is too low to handle fast human motion, resulting in large displacements between adjacent frames. Moreover, the 4D representation is unable to interpolate along the temporal dimension to recover plausible motion. As a result, the images rendered by the 4D Gaussian model exhibit poor visual quality (Fig.\ref{fig:ablation}(b)), and even the artifact-fix video diffusion model fails to generate satisfactory outputs (Fig.\ref{fig:ablation}(c)).
%
In contrast, with asynchronous capture (Fig.~\ref{fig:ablation}(d)), although some artifacts remain due to sparse views, the reconstructed motion is more accurate. The results are clearly superior to those from synchronous capture, further demonstrating the effectiveness of asynchronous acquisition for high-frame-rate 4D reconstruction in large-motion scenarios.

\subsection{Artifact-fix Video Diffusion Model} 
\textbf{Effectiveness of the artifact-fix video diffusion model. }
%
To demonstrate the effectiveness of our artifact-fix model, we compare rendering results with and without it. The comparison is presented in Table~\ref{tab:ablation} and Fig.\ref{fig:ablation}(d)(e). Leveraging the strong spatiotemporal priors of the video diffusion model, our approach effectively removes “floater” artifacts caused by sparse views in asynchronous capture, restores fine textures, and eliminates temporal discontinuities in the rendered video. By optimizing with the diffusion loss (Eqn.\ref{equa:diff}), we distill the diffusion prior into the 4D reconstruction pipeline, ultimately producing visually compelling results, as shown in Fig.~\ref{fig:ablation}(e).

\begin{figure}[!t]
   \centering
   \includegraphics[width=0.44\textwidth]{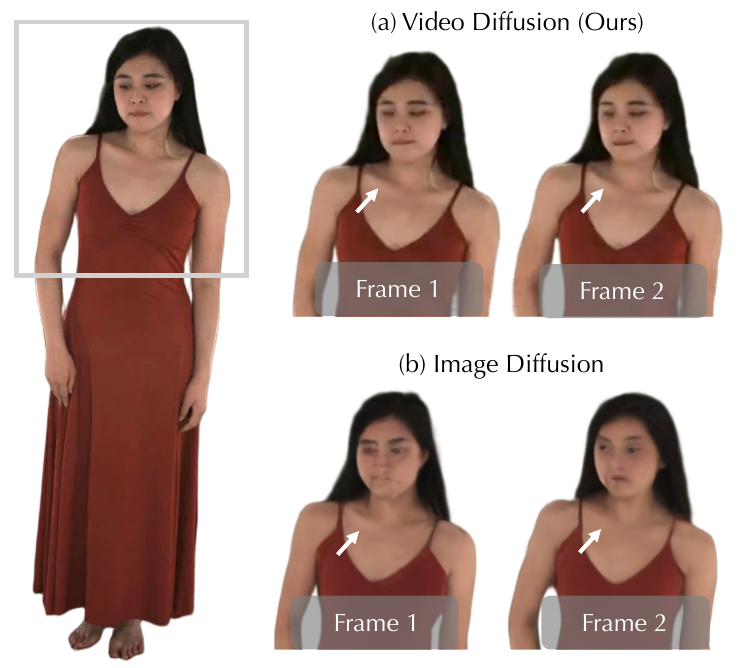}
   \caption{Ablation study on video diffusion model and image diffusion model.}
   \label{fig:sd}
\end{figure}

\noindent \textbf{Comparing video diffusion with image diffusion.}
%
To verify that using a video diffusion model as the backbone for the 4D artifact-fix task outperforms image-based diffusion, we train a Stable Diffusion 1.5 model with the ControlNet Tile module~\cite{zhang2023adding} on the same dataset, following~\cite{yu2024lm,yang2024gaussianobject}. The outputs of the two diffusion models are shown in Fig.~\ref{fig:sd}. We observe that, beyond differences in facial generation quality due to the base model's capabilities, the video diffusion model produces temporally consistent textures around the collarbone across adjacent frames. In contrast, the image diffusion model introduces noticeable texture variation in this region due to its frame-wise randomness. This temporal inconsistency poses a major challenge to maintaining smooth motion in 4D reconstruction.

\begin{figure}[!t]
   \centering
   \includegraphics[width=0.45\textwidth]{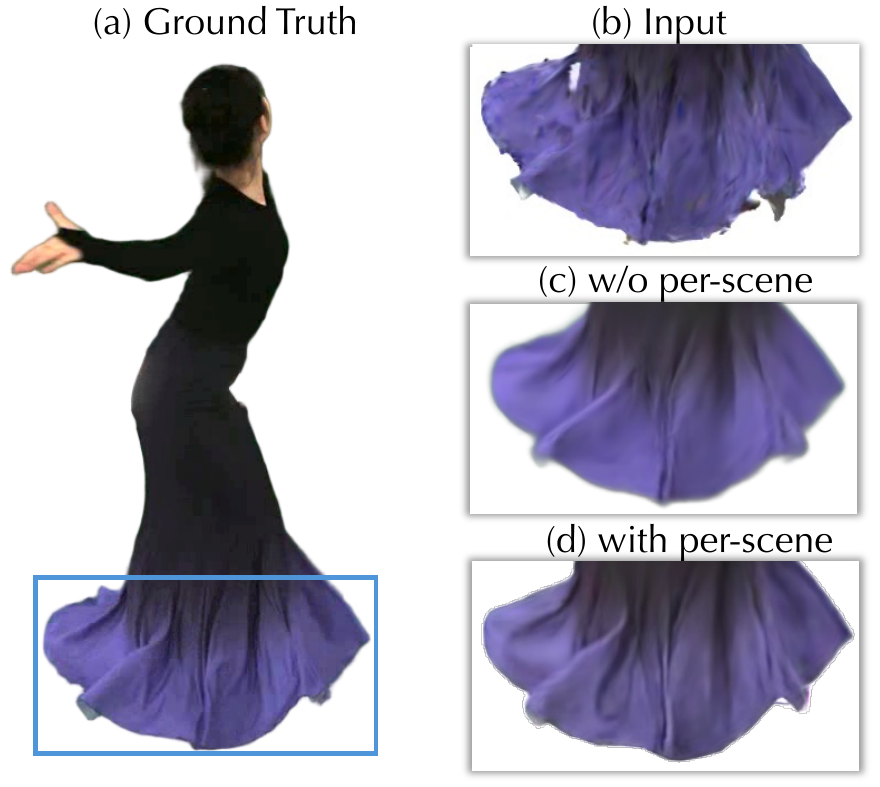}
   \caption{{Ablation study on per-scene finetuning.}}
   \label{fig:perscene}
\end{figure}

\subsection{Per-scene Finetuning Artifact-Fix Model}
To further improve visual quality, we also explored the idea of using per-scene optimization strategy for our artifact-fix video diffusion model. When the input to the artifact-fix video model is complex—such as scenes in Fig.~\ref{fig:perscene}(b) where some regions lack texture while others contain dense artifacts—the model may struggle to distinguish between noise that should be removed and texture details that should be preserved. Consequently, its performance may become suboptimal. To better incorporate scene-specific information, we explore the idea of per-scene finetune the artifact-fix model. Specifically, we adopt a leave-one-out strategy (similar to \cite{yang2024gaussianobject}) to construct noisy-clean video pairs from the input video, and then finetune the artifact-fix model on these pairs for a small number of iterations. The results in Fig.~\ref{fig:perscene}(c)(d) demonstrates that per-scene fine-tuning enables the model to recover finer dress details, as the diffusion model leverages multi-view and temporal cues to learn 4D scene distributions. 
While this strategy improves performance in complex scenes, the leave-one-out construction introduces significant time overhead. Therefore, \textbf{we do not adopt per-scene finetuning as the default setting in this work}. Nevertheless, we recognize it as a promising direction for high-quality reconstruction and include it in our discussion to inspire future research.

\subsection{Compare with frame interpolation methods}
To compare with interpolation methods, we first use state-of-the-art models to generate intermediate frames, and then train GS4D on the high-frame-rate videos. We benchmarked two baselines—CNN-based model RIFE~\cite{huang2022real} and diffusion-based model MoMo~\cite{lew2025disentangled}—on the DNA-Rendering Dataset, with the results shown in Table \ref{tab:result-inter}. For our target scenes with large motion, CNN-based models cause severe artifacts and deformation, while diffusion-based models generate incorrect motion. Therefore, neither performs well in these challenging scenarios.

\begin{table}[t] \setlength{\abovecaptionskip}{0.2cm}
\caption{\textbf{Quantitative comparison with frame interpolation methods.} Higher values indicate better performance for metrics marked with ($\uparrow$), while lower values are preferable for metrics marked with ($\downarrow$).}
    \centering
    {
    \begin{tabular}{lccccc}
    
    \hline
    \textbf{Method}  & \textbf{PSNR} $\uparrow$ &\textbf{SSIM } $\uparrow$ & \textbf{LPIPS} $\downarrow$ \\ \hline
    RIFE~\cite{huang2022real} &25.06&0.824&0.316\\       
           MoMo~\cite{lew2025disentangled} &24.85&0.811&0.324 \\
 \hline
      Ours & \textbf{26.76} & \textbf{0.845} & \textbf{0.293}\\
 \hline
    \end{tabular}}
    
     \label{tab:result-inter}
\end{table} 

\subsection{Limitation}
Due to the inherent fidelity limitations of pre-trained diffusion models, some fine textures, despite the per-scene optimization, may still exhibit reduced fidelity after artifact removal. Future advancements in foundational models could further enhance the reconstruction accuracy of asynchronous capture. In addition, our current artifact-fixing model is trained on the DNA-Rendering dataset, which primarily features human performances. Consequently, while our method demonstrates strong results on such scenes, its generalization to more diverse, in-the-wild scenarios with non-human subjects or different types of motion has not yet been verified. A promising direction for future work is to train our video diffusion model on a broader range of dynamic 4D datasets as they become available.

\section{Conclusion}

In conclusion, we have presented a novel asynchronous capture scheme and reconstruction pipeline to address the challenge of fast-dynamic scene reconstruction. By introducing deliberate delays between cameras, our method effectively increases the frame rate, enabling more accurate modeling of intermediate motion information in fast-moving scenes. 
To address the viewpoint sparsity and reconstruction artifacts introduced by asynchronous capturing, we train a video-diffusion-based artifact-fix model that significantly reduces these artifacts and enhances visual quality. Compared to previous sparse 3D reconstruction methods that employ image diffusion for refinement, our video diffusion–based approach effectively resolves the temporal inconsistency issues that arise in 4D reconstruction by leveraging spatiotemporal priors from video diffusion. Our method achieves superior performance in both real and synthetic scenarios, demonstrating its robustness across diverse conditions.
Furthermore, we provide the first real-world asynchronous multi-view video dataset to advance research in high-speed 4D reconstruction.
Our approach provides a promising joint hardware-software solution to improve the accuracy of 4D reconstruction for fast-motion scenes, with potential applications in sports, autonomous driving, robotics, and VR/AR content creation.

\section{Acknowledgements}
{
This work was supported by the RGC Early Career Scheme (ECS) No. 24209224. This study was also supported in part by the Centre for Perceptual and Interactive Intelligence (CPII) Ltd., a CUHK-led InnoCentre under the InnoHK initiative of the Innovation and Technology Commission of the Hong Kong Special Administrative Region Government.}


%
%
%
%

\bibliographystyle{ACM-Reference-Format}
\bibliography{sample-bibliography}



\appendix

\end{document}